\documentclass[a4paper,10pt]{article}
\pdfoutput=1
\usepackage[bookmarks=false]{hyperref}
\usepackage[numbers,sort&compress]{natbib}
\usepackage[margin=1in]{geometry}

\usepackage{cite}

\usepackage[cmex10]{amsmath}
  \usepackage[ruled, vlined, nofillcomment]{algorithm2e}
  \usepackage{algorithmic}
  
  \SetCommentSty{mycommfont}
  \SetAlgoCaptionSeparator{:}
  
\usepackage{calc}

\makeatletter
\let\MYcaption\@makecaption
\makeatother

\usepackage[font=footnotesize]{subcaption}

\makeatletter
\let\@makecaption\MYcaption
\makeatother

\newcommand{\eg}{\textit{e.g.}}
\newcommand{\ie}{\textit{i.e.}}

\usepackage[T1]{fontenc}
\usepackage[utf8]{inputenc}

\usepackage{amsmath,amsfonts,amssymb,amsthm,mathrsfs}

\usepackage{bm}
\newcommand{\mathbold}[1]{\bm{#1}}
\newcommand{\mbf}[1]{\mathbf{#1}}
\newcommand{\vect}[1]{\mbf{#1}}
\newcommand{\vectb}[1]{\bm{#1}}
\newcommand{\mat}[1]{\vect{#1}}

\newcommand{\T}{^\mathsf{T}}    % Transpose
\newcommand{\R}{\mathbb{R}}    % Real numbers
\newcommand{\N}{\mathcal{N}}   % Gaussian distribution

\DeclareMathOperator{\diag}{diag}

\newcommand{\vepsilon}[0]{\mathbold{\varepsilon}}

\newcommand{\wframe}{\text{w}}
\newcommand{\rotwb}{\text{wb}}
\newcommand{\rotbw}{\text{bw}}
\newcommand{\lin}{\text{l}}
\newcommand{\nlin}{\text{n}}
   
\usepackage{color}
\usepackage{tikz,pgfplots}
\usetikzlibrary{plotmarks}

\newlength{\figureheight}

\usepackage{framed}

\usepackage{url}
\begin{document}
% !TEX spellcheck = en-UK

\newcommand{\coverTitle}{Scalable Magnetic Field SLAM in 3D Using Gaussian Process Maps}
\newcommand{\coverYear}{2018}

\newcommand{\coverAuthors}{Manon~Kok$^{\star,1}$ and Arno~Solin$^{\dagger,1}$ \\ \vspace{3mm}
\small{$^\star$Delft University of Technology, the Netherlands, e-mail: m.kok-1@tudelft.nl} \\
\small{$^\dagger$Aalto University, Finland, e-mail: arno.solin@aalto.fi} 
}

\footnotetext[1]{This work was undertaken whilst MK was a Research Associate at the Department of Engineering, University of Cambridge, and AS was a Visiting Research Fellow at the Department of Engineering, University of Cambridge.}

\begin{titlepage}
\begin{center}
%
%% 
%{\large \em Technical report}

\vspace*{2.5cm}
%
%% TITLE
{\Huge \bfseries \coverTitle  \\[0.4cm]}

%
%% AUTHORS
{\Large \coverAuthors \\[1.5cm]}

\renewcommand\labelitemi{\color{red}\large$\bullet$}
\begin{itemize}
\item {\Large \textbf{Please cite this version:}} \\[0.4cm]
\normalsize
Manon Kok and Arno Solin (2018), "Scalable Magnetic Field SLAM in 3D Using Gaussian Process Maps", Proceedings of the 20th International Conference on Information Fusion, Cambridge, UK, July 2018.
\end{itemize}

\end{center}

\vspace{1cm}

\begin{abstract}
\noindent We present a method for scalable and fully 3D magnetic field simultaneous localisation and mapping (SLAM) using local anomalies in the magnetic field as a source of position information. These anomalies are due to the presence of ferromagnetic material in the structure of buildings and in objects such as furniture. We represent the magnetic field map using a Gaussian process model and take well-known physical properties of the magnetic field into account. We build local maps using three-dimensional hexagonal block tiling. To make our approach computationally tractable we use reduced-rank Gaussian process regression in combination with a Rao--Blackwellised particle filter. We show that it is possible to obtain accurate position and orientation estimates using measurements from a smartphone, and that our approach provides a scalable magnetic field SLAM algorithm in terms of both computational complexity and map storage.
\end{abstract}

\vfill

\end{titlepage}

\title{Scalable Magnetic Field SLAM in 3D \\ Using Gaussian Process Maps}

\author{Manon~Kok$^\star$ and Arno Solin$^\dagger$ \\
\small{$^\star$Delft University of Technology, the Netherlands, e-mail: m.kok-1@tudelft.nl} \\
\small{$^\dagger$Aalto University, Finland, e-mail: arno.solin@aalto.fi} \\
}
\date{\empty}

%\maketitle
%
%\begin{abstract}
%  We present a method for scalable and fully 3D magnetic field simultaneous localisation and mapping (SLAM) using local anomalies in the magnetic field as a source of position information. These anomalies are due to the presence of ferromagnetic material in the structure of buildings and in objects such as furniture. We represent the magnetic field map using a Gaussian process model and take well-known physical properties of the magnetic field into account. We build local maps using three-dimensional hexagonal block tiling. To make our approach computationally tractable we use reduced-rank Gaussian process regression in combination with a Rao--Blackwellised particle filter. We show that it is possible to obtain accurate position and orientation estimates using measurements from a smartphone, and that our approach provides a scalable magnetic field SLAM algorithm in terms of both computational complexity and map storage.
%\end{abstract}

\section{Introduction}
\label{sec:introduction}
\noindent
The use of the magnetic field as a source of position information for indoor navigation is a promising novel approach that has gained interest in recent years~\citep{Haverinen+Kemppainen:2009,Li+Gallagher+Dempster+Rizos:2012,Angermann+Frassl+Doniec+Julian+Robertson:2012,LeGrand+Thrun:2012,Solin+Sarkka+Kannala+Rahtu:2016,Hanley+Faustino+Zelman+Degenhardt+Bretl:2017,Gao+Harle:2017,Torres+Rambla+Montoliu+Belmonte+Huerta:2015}. It relies upon the spatial variation of the ambient magnetic field which is typically due to ferromagnetic material in the structures of buildings and to a lesser extent due to the presence of \eg~furniture. The advantage of using the magnetic field for positioning is that it can be measured by a small device, without additional infrastructure and without line-of-sight requirements. Furthermore, magnetometers are nowadays present in (almost) any smartphone. Crucial to this approach is the ability to build maps of the magnetic field that can be used for interpolation and extrapolation and the ability to localise inside this map. In this paper, we present a scalable and fully three-dimensional \emph{simultaneous localisation and mapping (SLAM)} approach that builds a map of the magnetic field while simultaneously localising the sensor in the map.

\begin{figure}[!t]
  \includegraphics[width=\columnwidth]{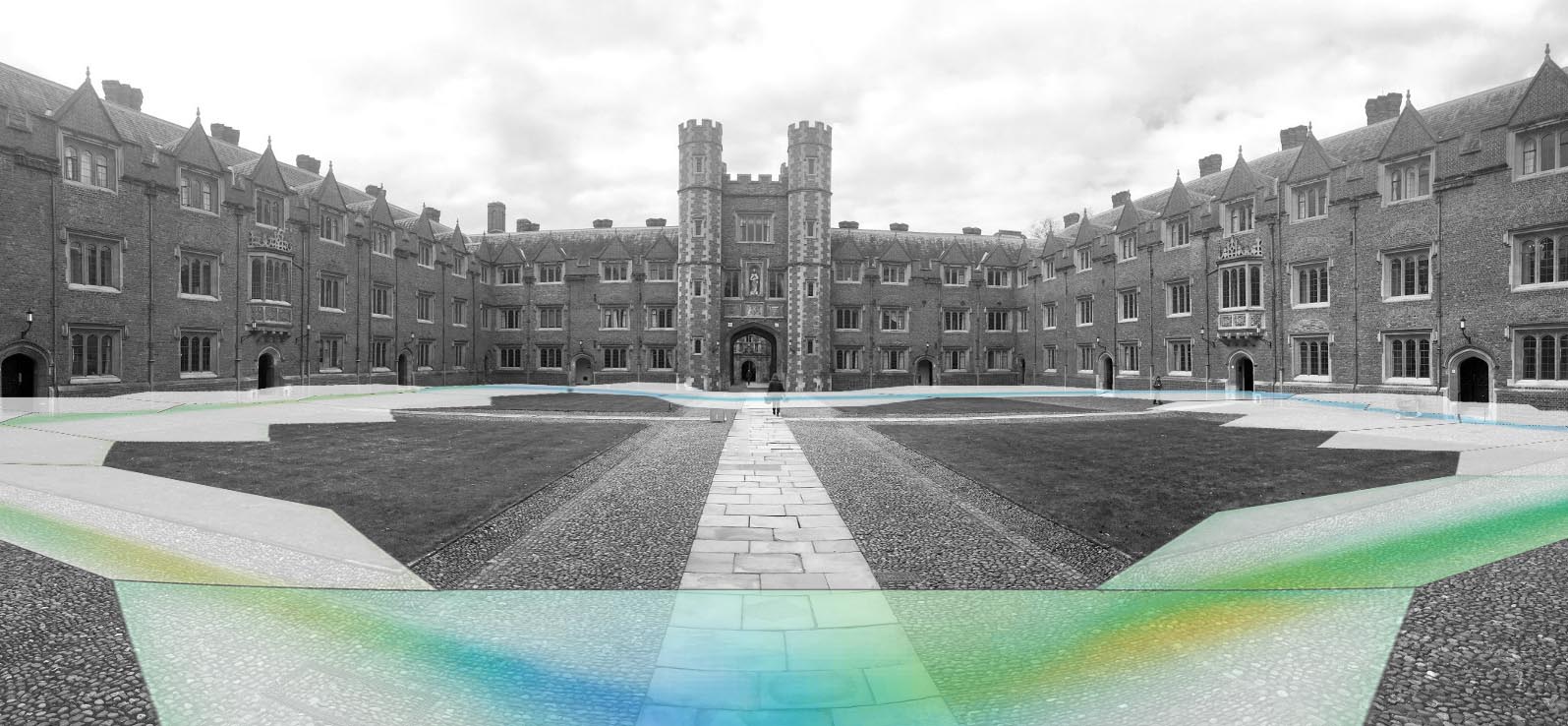}
  \caption{Illustration of a magnetic field map built using our SLAM approach. Each hexagonal block (with radius of 5~m and height of 4~m) covers a Gaussian process magnetic map volume of $\sim$260~$\text{m}^3$ in a state dimension of 256. A more detailed map is provided in Fig.~\ref{fig:st-johns}.}
  \label{fig:intro}
\end{figure}

Extending previous work~\citep{Solin+Kok+Wahlstrom+Schon+Sarkka:2015}, we build the magnetic field map using Gaussian process regression~\citep{Rasmussen+Williams:2006} and incorporate physical knowledge about the magnetic field known from Maxwell's equations in the Gaussian process prior. To overcome issues with computational complexity, we use the rank-reduced approach first presented in~\citep{Solin+Sarkka:submitted}. This results in a representation that fits perfectly in a \emph{Rao--Blackwellised particle filter (RBPF)}~\citep{Schon+Gustafsson+Nordlund:2005} which can be used both for building the map and for localisation in the map, resulting in a tractable localisation algorithm. Example results can be found in Fig.~\ref{fig:intro}. 

The approach presented in~\citep{Solin+Sarkka:submitted} relies on a basis function expansion on a specific domain where the required number of basis functions scales with the size of the domain. To allow for mapping large 3D areas, we propose a map representation using hexagonal block tiling, which aims at providing a very compact (in terms of required storage per volume) representation of the magnetic field map (all three vector field components and associated uncertainties). Using this approach, hexagonal block tiles are created whenever particles enter unexplored areas, resulting in a growing number of local Gaussian process maps.

We present experimental results where we obtain magnetometer measurements as well as odometry from a smartphone. This odometry gives us information about the change in position and orientation but the resulting pose estimates drift over time. This drift is corrected by making use of the magnetometer measurements and the magnetic field map. We show that accurate 6D pose and 3D map estimates can be obtained in large scale and 3D experiments, as illustrated in Fig.~\ref{fig:intro}.

Previous work on magnetic field SLAM typically assumes that the motion is in 2D or 2.5D \citep{Robertson+Frassl+Angermann+Doniec+Julian+Puyol+Khider+Lichtenstern+Bruno:2013,Frassl+Angermann+Lichtenstern+Robertson+Julian+Doniec:2013,Vallivaara+Haverinen+Kemppainen+Roning:2010, Vallivaara+Haverinen+Kemppainen+Roning:2011}. To the best of the authors' knowledge, this is the first fully 3D approach to magnetic field SLAM. It is also the first time the reduced-rank approach from~\citep{Solin+Sarkka:submitted} has been used for SLAM and that the domain in this approach has been chosen to be hexagonal. 

\section{Problem formulation}
\label{sec:probForm}
\noindent
We are interested both in estimating the position and the orientation of the sensor and in building a map of the magnetic field. We denote the position of the sensor at time instance $t$ by $\vect{p}_t^\wframe$. The superscript $\wframe$ refers to the \emph{world frame}. In our SLAM formulation, it is always possible to move or rotate the estimated map while simultaneously moving and rotating the position and orientation estimates. In other words, the absolute position and orientation are not observable. Because of this, without loss of generality we choose the origin of the world frame to be equal to the initial position of the sensor and its axes to be aligned with the coordinate system defined by the initial orientation. 

The orientation of the sensor is denoted by $\vect{q}^\rotwb_t$, where~$\text{b}$ refers to the \emph{body frame}. Its origin lies in the centre of the magnetometer triad and its axes are aligned with the axes of the magnetometer. We encode the orientation using a unit quaternion. There is a direct mapping from the quaternion~$\vect{q}^\rotwb_t$ to the rotation matrix $\vect{R}^\rotwb_t$ and because of this we use these interchangeably. The double superscript $\rotwb$ denotes the rotation from body frame to world frame. The reverse rotation is denoted by $\rotbw$ and hence, $\vect{R}^\rotwb_t = \left( \vect{R}^\rotbw_t \right)\T$.

We denote our estimate of the magnetic field map by~$\vect{m}_t$. More details about the exact form of $\vect{m}_t$ can be found in Section~\ref{sec:gp-map}. Note that we assume that the map of the magnetic field does not change over time. The subscript $t$ refers to the fact that our belief of the map changes over time because more data results in more information about the map. 

The resulting state vector $\vect{x}_t$ is given by
\begin{align}
\vect{x}_t = \begin{pmatrix} (\vect{p}_t^\wframe)\T & (\vect{q}^\rotwb_t)\T & \vect{m}_t\T \end{pmatrix}\T.
\label{eq:stateVector}
\end{align}
The dynamics of the position and orientation are modelled in terms of the change in position $\Delta \vect{p}_t$ and orientation $\Delta \vect{q}_t$ obtained from the odometry information. We consider these to be inputs to our dynamic model. Using the assumption that the magnetic field does not vary over time, our dynamic model can be written as 
\begin{align}
\vect{x}_{t+1} &= \begin{pmatrix} \vect{p}_t^\wframe + \Delta \vect{p}_t + \vepsilon_{\text{p},t} \\ \Delta \vect{q}_t \odot \vect{q}^\rotwb_t \odot \exp_\text{q} (\vepsilon_{\text{q},t}) \\ \vect{m}_t  \end{pmatrix}.
\label{eq:dynModel}
\end{align}
We assume that the noise on the position ($\vepsilon_{\text{p},t}$) and on the orientation ($\vepsilon_{\text{q},t}$) is Gaussian with $\vepsilon_{\text{p},t} \sim \N(\vect{0}, \Delta T \sigma_{\text{p}}^2 \, \mat{I}_3)$ and $\vepsilon_{\text{q},t} \sim \N(\vect{0}, \Delta T \sigma_{\text{q}}^2 \, \mat{I}_3)$. Here, $\Delta T$ is the (possibly time-varying) time between subsequent samples. In~\eqref{eq:dynModel}, $\odot$ denotes the quaternion multiplication and $\exp_\text{q}(\vepsilon_{\text{q},t})$ refers to the representation of the noise vector $\vepsilon_{\text{q},t} \in \mathbb{R}^3$ as a unit quaternion. The dynamics of the orientation $\vect{q}^\rotwb_t$ can be interpreted as the rotation equivalent of the dynamics of the position $\vect{p}_t^\wframe$. For more information about quaternion algebra and representations of orientation, see \eg \citep{Kok+Hol+Schon:2017}.

We use the magnetometer measurements to provide information about the magnetic field map as well as the position and orientation of the sensor. This can be modelled as
\begin{align}
\label{eq:measModel}
\vect{y}_{\text{m},t} &= \vect{R}^\rotbw_t \nabla \vectb{\Phi}_t(\vect{p}_t^\wframe) \vect{m}_t + \vepsilon_{\text{m},t}. 
\end{align}
Here, $\nabla \vectb{\Phi}_t(\vect{p}_t^\wframe) \vect{m}_t$ is the prediction of the magnetic field at the location $\vect{p}_t^\wframe$. More details about the specific form of $\nabla \vectb{\Phi}_t(\vect{p}_t^\wframe)$ can be found in Section~\ref{sec:gp-map}. Note that the magnetic field map is represented in world frame. Hence, the magnetometer measurements are modelled as the prediction $\nabla \vectb{\Phi}_t(\vect{p}_t^\wframe) \vect{m}_t$, rotated to body frame.

Together with a prior on the state at the first time instance, the models~\eqref{eq:dynModel} and~\eqref{eq:measModel} constitute the state space model that we use in the RBPF to estimate the state defined in~\eqref{eq:stateVector}. Because the initial position and orientation fix the world coordinate frame $\text{w}$, $\vect{p}^\wframe_1$ and $\vect{q}^\rotwb_1$ are set to the initial position and orientation with zero uncertainty. The magnetic field map at the first time instance is set to be equal to the Gaussian process prior. More details about the Gaussian process map will be discussed in Section~\ref{sec:gp-map}. The RBPF implementation will subsequently be discussed in Section~\ref{sec:rbpf}.

\section{Methods}

\subsection{Gaussian process magnetic field map}
\label{sec:gp-map}
\noindent
Building on previous work~\citep{Solin+Kok+Wahlstrom+Schon+Sarkka:2015}, we build a Gaussian process magnetic field map in which we encode physical knowledge about the magnetic field. First of all, we know that the measured magnetic field consists of the Earth's magnetic field and the magnetic field due to anomalies induced by small-scale variations and building structures. Furthermore, using Maxwell's equations and classical electromagnetism~\citep{Griffiths+College:1999}, we know that the magnetic field due to the anomalies can be modelled using a \emph{latent scalar potential field} $\varphi(\vect{p})$. Here, $\varphi : \R^3 \to \R$, where $\vect{p} \in \R^3$ is the spatial coordinate in world frame and the magnetometer measures the \emph{derivative} of this scalar potential. We assume the scalar potential field to be a realisation of a Gaussian process prior and the magnetometer measurements to be corrupted by Gaussian noise, resulting in the following model
\begin{equation}
\begin{split} \label{eq:scalar-potential-gp}
\varphi(\vect{p}) &\sim \mathrm{GP}(0,\kappa_\text{lin.}(\vect{p},\vect{p}') + \kappa_\text{SE}(\vect{p},\vect{p}')), \\
\vect{y}_{\text{m},t} &= \vect{R}_t^\rotbw \nabla \varphi(\vect{p}) \big|_{\vect{p}=\vect{p}^\wframe_t} + \vectb{\varepsilon}_{\text{m},t},
\end{split}
\end{equation}
where $\vectb{\varepsilon}_{\text{m},t} \sim \N(\vect{0}, \sigma_\text{m}^2 \, \mat{I}_3)$, for each observation $t=1,2,\ldots,N_T$. The local Earth's magnetic field contributes \emph{linearly} to the scalar potential which we model in terms of a linear covariance function~\citep{Duvenaud:2014} as
\begin{equation} \label{eq:cf-linear}
\kappa_\text{lin.}(\vect{p},\vect{p}') = 
\sigma_\text{lin.}^2 \, 
\vect{p}\T\vect{p}',
\end{equation}
where $\sigma_\text{lin.}^2$ is the magnitude scale hyperparameter. The magnetic field anomalies are captured by the squared exponential covariance function
\begin{equation} \label{eq:cf-SE}
\kappa_\text{SE}(\vect{p},\vect{p}') = \sigma_\text{SE}^2 \, \exp\left(-\frac{\|\vect{p}-\vect{p}'\|^2}{2\ell^2}\right),
\end{equation}
where $\sigma_\text{SE}^2$ is a magnitude and $\ell$ the characteristic length-scale hyperparameter. Including the additional prior information introduced by the scalar potential in~\eqref{eq:scalar-potential-gp} results in a coupling between the three components of the magnetic field according to physical laws. This has been shown in \citep{Solin+Kok+Wahlstrom+Schon+Sarkka:2015} to improve prediction accuracy over simply modelling each magnetic field component separately. 

Using the model~\eqref{eq:scalar-potential-gp}, it is possible to predict the magnetic field at previously unseen locations. In practice, however, this quickly becomes computationally intractable because of the large amount of magnetometer measurements we collect and the fact that the computational complexity scales as $\mathcal{O}(N_T^3)$. A computationally tractable solution was presented in \citep{Solin+Kok+Wahlstrom+Schon+Sarkka:2015} where the problem was projected on the eigenbasis of the (negative) Laplace operator in a confined domain $\Omega \subset \R^3$. In this domain, the eigendecomposition of the Laplace operator subject to Dirichlet boundary conditions can be solved as
\begin{equation}
\begin{cases}\label{eq:eigenbasis}
-\nabla^2 \phi_j(\vect{p}) = \lambda_j^2 \phi_j(\vect{p}), 
& \vect{p} \in \Omega, \\
\phantom{-\nabla^2} \phi_j(\vect{p}) = 0, 
& \vect{p} \in \partial\Omega.
\end{cases}
\end{equation}
Using this eigendecomposition, it is possible to approximate the covariance function in~\eqref{eq:scalar-potential-gp} as
\begin{align}
\kappa(\vect{p},\vect{p}') & =  \kappa_\text{lin.}(\vect{p},\vect{p}') + \kappa_\text{SE}(\vect{p},\vect{p}') \notag \\
& \approx \kappa_\text{lin.}(\vect{p},\vect{p}') + \sum_{j=1}^m S_\text{SE}(\lambda_j) \, \phi_j(\vect{p})\,\phi_j(\vect{p}'),
\label{eq:approxKernel}
\end{align}
where $\kappa_\text{SE}(\vect{p},\vect{p}') $ is approximated using $m$ basis functions $\phi_j$ and their corresponding eigenvalues $\lambda_j$. The exact form of $\phi_j$ and $\lambda_j$ depends on the shape of the domain $\Omega$. Note that $S_\text{SE}(\cdot)$ denotes the spectral density function of the squared exponential kernel $S_\text{SE}(\cdot)$ for which a closed form solution exists, see~\citep{Solin+Kok+Wahlstrom+Schon+Sarkka:2015}. Using the approximation~\eqref{eq:approxKernel} with the boundary conditions defined in~\eqref{eq:eigenbasis} implies that our model of the magnetic field reverts back to the Earth's magnetic field at the boundary of the domain. 

A fixed domain $\Omega$ is not desirable in SLAM, as we do not know {\it a~priori} the spatial extent of the final magnetic field map. Furthermore, the number of basis functions that is needed for a good approximation scales with the size of the domain and for large scale SLAM problems will hence become intractable. We therefore propose a scalable representation where we split the magnetic map into a three-dimensional grid of hexagonal blocks with radius $r$ and height $2 L_\mathrm{z}$, such that each subdomain~$d$ is given by
\begin{align}
\label{eq:hex}
  \Omega^{(d)} = \{ \vect{p} \mid (p_1,p_2) \in \mathrm{hexagon}(r,\vect{p}_{1,2}^{(d)}), \qquad p_3 \in [{p}_{3}^{(d)}-L_\mathrm{z}, {p}_{3}^{(d)}+L_\mathrm{z}] \}.
\end{align}
Here, $\vect{p}^{(d)}$ is the center point of the $d$th hexagon cell in the grid representation. The hexagonal grid provides an efficient way of maximising the area per number of basis functions for representing the magnetic field map, thus requiring less memory for storing the map representation.

\begin{figure}[!t]
  \centering
%  \pgfplotsset{yticklabel style={rotate=90}, ylabel style={yshift=-15pt},clip=true,scale only axis}
%  \setlength{\figurewidth}{\columnwidth}
%  %
%  \input{fig/hexagon_eigenbasis.tex}
  \includegraphics[width=0.7\columnwidth]{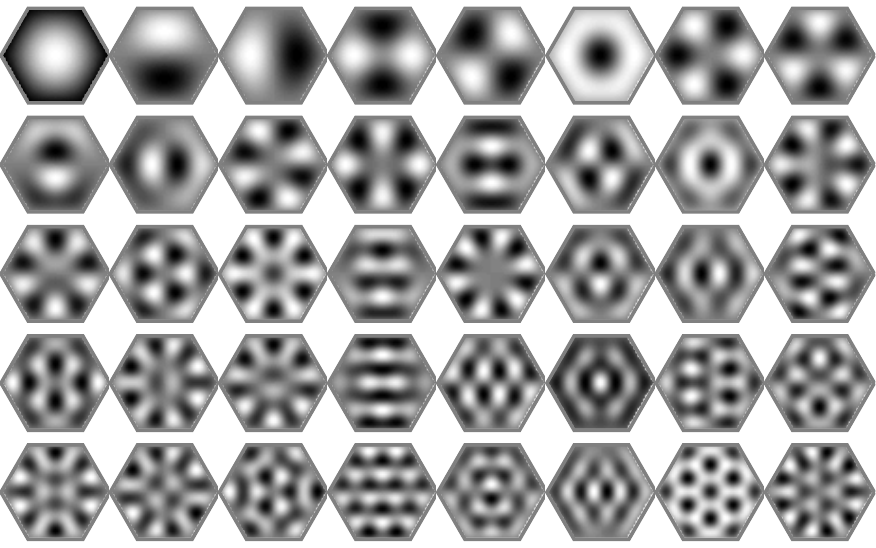}
  \caption{The first 40 eigenfunctions of the negative Laplace operator in a hexagonal domain subject to Dirichlet boundary conditions. The eigenbasis can be effectively solved by a numerical solver, and the calculations are only required once.}
  \label{fig:hexagon-eigenbasis}
\end{figure}

To compute the basis functions $\phi_j$ and their corresponding eigenvalues $\lambda_j$, we need to solve the eigendecomposition of the Laplace operator~\eqref{eq:eigenbasis} in a two-dimensional hexagon. This can not be solved in closed form, but instead it can be solved numerically. We set up the eigenvalue problem by composing the sparse stencil matrix corresponding to the Laplacian using a nine-point finite difference approximation. The solution itself is given by the Lanczos algorithm (see, \eg, \citep{Golub+VanLoan:2012}, callable in Matlab through \texttt{eigs}) that we use for solving the $m$ largest real eigenvalues $[\lambda_{n}^\mathrm{hex}]^2$---together with the corresponding eigenfunctions $\phi_{n}^\mathrm{hex}(\vect{p})$. Fig.~\ref{fig:hexagon-eigenbasis} shows the 40 first eigenfunctions of the negative Laplacian in a unit hexagonal domain with respect to Dirichlet boundary conditions. 

The eigendecomposition is extended to cover the three-dimensional hexagonal grid cell $\Omega^{(d)}$ by considering the numerically solved horizontal eigendecomposition and a closed-form solution to the separable vertical dimension. This choice is motivated by the assumption that paths typically exhibit most variation in the horizontal direction while exhibiting less variation in the vertical direction. In other words, we assume most variation to happen in horizontal cross sections since human motion tends to be along two-dimensional planes with less-frequent ascending and descending. None of the computational benefits rely on this assumption, and it is also a sensible choice for other (non-bipedal) movement.

The final eigenbasis then becomes the combination:
\begin{align}
  \phi_j(\vect{p}) &= 
    \phi_{n_{j,1}}^\mathrm{hex}(\vect{p}_{1,2}) \, 
    \left[\frac{1}{\sqrt{L_\mathrm{z}}} \sin\!\bigg( \frac{\pi n_{j,2} (p_3 + L_\mathrm{z})}{2 L_\mathrm{z}} \bigg)\right], \label{eq:basisFunctions}\\
  \lambda_j^2 &=
    [\lambda_{n_{j,1}}^\mathrm{hex}]^2 +
    \bigg( \frac{\pi n_{j,2}}{2 L_d} \bigg)^2,
\end{align}
where the matrix $\vect{n} \in \R^{m \times 2}$ consists of an index set of permutations of integers $\{1,2,\ldots,m\}$ (\ie, $\{(1, 1),(2,1),\ldots,(1,2),\ldots,(2,2), \ldots\}$) corresponding to the $m$ largest eigenvalues. 

The numerical solution for the two-dimensional hexagon is fast and takes less than a second even with a tight discretisation on an average laptop computer. Furthermore, because the basis functions~\eqref{eq:basisFunctions} are independent of the hyperparameters and the input locations, they only need to be evaluated once, not causing any computational overhead in our SLAM algorithm.

The basis function expansion described above allows us to write the prior model for each hexagonal tile at time $t=0$ in terms of a mean $\vect{m}_0 = \vect{0}_{m+3}$ and a covariance 
\begin{equation}\label{eq:basis-Lambda}
    \vect{P}_0 = \diag \left( {\sigma_\text{lin.}^2, \sigma_\text{lin.}^2, \sigma_\text{lin.}^2, S_\text{SE}(\lambda_1), S_\text{SE}(\lambda_2), \ldots, S_\text{SE}(\lambda_m)} \right).
\end{equation}
The posterior can be updated sequentially~\citep{Sarkka:2013} as new magnetometer data arrives at $t=1,2,\ldots,N_T$ and can be written as
\begin{align}\label{eq:Kalman-update}
\begin{split}
\vect{S}_t       &= 
\vect{C}_t \vect{P}_{t-1} \vect{C}_t\T +
\sigma_\text{m}^2 \, \vect{I}_3, \\
\vect{K}_t       &= 
\vect{P}_{t-1} \vect{C}_t\T \vect{S}_t^{-1}, \\
\vect{m}_t    &= 
\vect{m}_{t-1} + \vect{K}_t (\vect{y}_{\text{m},t} - \vect{C}_t \, \vect{m}_{t-1}), \\
\vect{P}_t &= 
\vect{P}_{t-1} - \vect{K}_t \vect{S}_t \vect{K}_t\T, 
\end{split}
\end{align}
with $\vect{C}_t = \vect{R}_t^\rotbw \nabla \vectb{\Phi}_t$ and
\begin{equation}\label{eq:basis-dx}
  \nabla \vectb{\Phi}_t= 
  \begin{pmatrix}
    \nabla \vect{p}_t\T, \nabla\phi_1(\vect{p}_t), \nabla\phi_2(\vect{p}_t), \ldots, \nabla\phi_m(\vect{p}_t)
  \end{pmatrix}.
\end{equation}
Note that~\eqref{eq:Kalman-update} is a Kalman filter measurement update of the state $\vect{m}_t$ representing the magnetic field map in a specific tile. This fact will be exploited in our magnetic field SLAM algorithm presented in Section~\ref{sec:rbpf}. 

\subsection{Magnetic field SLAM}
\label{sec:rbpf}
\noindent
As discussed in Section~\ref{sec:probForm}, we are interested in estimating the position and the orientation of the sensor while building a map of the magnetic field. We represent the map in terms of a varying number of hexagonal block tiles, each using the basis function expansion presented in Section~\ref{sec:gp-map}. We use a nonlinear filtering technique to estimate the position, the orientation and the map of each tile that has been visited. The state vector~\eqref{eq:stateVector} has a relatively large dimension due to the representation of the magnetic field map $\vect{m}_t$. For instance, in Section~\ref{sec:results} we set the dimension of the state $\vect{m}_t$ to $256$ per hexagonal block. However, $\vect{m}_t$ enters \emph{conditionally linearly} into the state space model. It is possible to exploit this by using an RBPF~\citep{Schon+Gustafsson+Nordlund:2005} which uses a Kalman filter for the conditionally linear states $\vect{x}_t^\lin$ and a particle filter for the nonlinear states $\vect{x}_t^\nlin$ where
\begin{align}
\vect{x}_t^\nlin = \begin{pmatrix} (\vect{p}_t^\wframe)\T & (\vect{q}^\rotwb_t)\T \end{pmatrix}, \qquad \vect{x}_t^\lin = \{ \vect{m}_t \}_{d = 1}^{N_{D,t}}.
\end{align}
Here, $N_{D,t}$ is the number of hexagonal block tiles that have been created up to time $t$. 

In the RBPF, $i = 1, \hdots, N_P$ particles having a position $\vect{p}_t^{\wframe,(i)}$ and an orientation $\vect{q}_t^{\rotwb,(i)}$ are used to represent the state at time $t$. Each of the particles also contains an estimate of the map $\vect{m}_t^{(i)}$ and its covariance $\vect{P}_t^{(i)}$. Note that $\vect{m}_t^{(i)}$ consists of the magnetic field map of each hexagonal block that particle~$i$ has visited. We assume that each tile is independent. The covariance of the linear state can therefore be described as $\vect{P}_t = \{ \vect{P}_t \}_{d = 1}^{N_{D,t}}$. 

The dynamics of the state is chosen as in~\eqref{eq:dynModel} and the measurement model is given in~\eqref{eq:measModel}. It is important to note that, as discussed in Section~\ref{sec:probForm}, the absolute position and orientation are not observable. Because of this, we set the initial position $\vect{p}^{\wframe,(i)}_0$ and orientation $\vect{q}^{\rotwb,(i)}_0$ of all particles $i = 1, \hdots, N_P$ equal to an initial position and orientation $\vect{p}_0^{\wframe,(i)} = \vect{0}_3$, $\vect{q}^{\rotwb,(i)}_0 = \vect{q}^{\rotwb}_0$. We initialise the magnetic field according to the Gaussian process prior (see also Section~\ref{sec:gp-map}).

The state space model presented in Section~\ref{sec:probForm} results in a fairly straightforward RBPF implementation. The major difference from a standard RBPF is that each particle only updates the local magnetic field map at its current location. Note that the computational complexity of the algorithm scales linearly with the number of particles and linearly with the number of hexagonal tiles per particle. The algorithm is summarised in Alg.~\ref{alg:rbpf}.

In a practical magnetic field SLAM implementation care must be taken of the following in terms of updating the map:
\begin{itemize}
\item To avoid boundary effects due to the Dirichlet boundary conditions, the actual tile extends slightly outside of the hexagonal block domain for updating. 
\item To allow for a smooth transition from one tile to the next, particles close to the border of a tile also update the map of the neighbouring tiles. 
\end{itemize}
During the ``exploration phase'' when particles start building a map of previously unseen locations, very little information is available to distinguish between the particles. Because of this, we would like to ensure that the particle cloud keeps spreading during this phase. To this end, we adapt the RBPF implementation in two ways:  
\begin{itemize}
\item To maintain the particle spread during the exploration phase, we delay the updating of the map by one lengthscale~$\ell$. 
\item To not collapse the particle cloud too quickly, we only resample the particles when they revisit a tile. Resampling is further delayed until at least $90\%$ of the particles has arrived in a tile that they revisit.
\end{itemize}

\begin{algorithm}[t]
  \caption{Magnetic field SLAM}
  \label{alg:rbpf}
  \begin{minipage}{\linewidth-14.45pt}
  \begin{algorithmic}[1]
    \REQUIRE $\{ \vect{y}_{\text{m},t}, \Delta \vect{p}_t^\wframe, \Delta \vect{q}_t \}_{t = 1}^{N_T}$, $\vect{q}^{\rotwb}_0$, $\Sigma_\text{p}$, $\Sigma_\text{q}$, $m$.
    \ENSURE For each $t$, $\vect{p}_t^\wframe$, $\vect{q}_t^\rotwb$, $\vect{m}_t$ of the highest-weight particle at $t$.
    \STATE \textit{Initialisation}: Initialise the particles as $\vect{p}_0^{\wframe,(i)} = \vect{0}_3$, $\vect{q}^{\rotwb,(i)}_0 = \vect{q}^{\rotwb}_0$, set $\vect{m}^{(i)}_0 = \vect{0}_{m+3}$, $\vect{P}^{(i)}_0$ as given in~\eqref{eq:basis-Lambda} and initialise the weights of the particles as $w_0^{(i)} = \tfrac{1}{N_P}$, for $i = 1, \hdots N_P$.
    \STATE \textbf{For} $t = 1, \hdots, N_T$, 
    \STATE \textit{Create new hexagonal block tiles:} If $\vect{p}_t^{\wframe,(i)}$ is in a new hexagonal block tile $d$, create this tile for particle $i$ and initialise $\vect{m}^{(i)}_{d,t} = \vect{0}_{m+3}$ and $\vect{P}^{(i)}_{d,t}$ as given in~\eqref{eq:basis-Lambda}.
    \STATE \textit{Evaluate the importance weights:} For $i = 1, \hdots, N_P$, evaluate the importance weights $w_t^{(i)}$ based on the measurement model~\eqref{eq:measModel} and normalise.
    \STATE \textit{Resampling}: If $90\%$ of the particles is in a tile that they are revisiting, resample those particles with replacement.
    \STATE \textit{Kalman filter measurement update:} 
    For $i = 1, \hdots, N_P$, update the magnetic field map $\vect{m}_{d,t}^{(i)}$ and its covariance $\vect{P}_{d,t}^{(i)}$ of the hexagon $d$ that the particle is in using the nonlinear states $\vect{p}_t^{\wframe,(i)}$ and $\vect{R}_t^{\rotwb,(i)}$ according to~\eqref{eq:Kalman-update}. If the particle is close to the border of a tile, also update the map of the neighbouring tiles.
	\STATE \textit{Obtain a point estimate:} Output the position, orientation, and estimated map of the particle with the highest weight.
        \STATE \textit{Particle filter time update:} For $i = 1, \hdots, N_P$, predict new particles $\vect{x}^{\nlin,(i)}_{t+1}$ as 
		\begin{align}
                    \vect{x}^{\nlin,(i)}_{t+1} &= \begin{pmatrix} \vect{p}_t^{\wframe,(i)} + \Delta \vect{p}_t^{\wframe,(i)} + 	\vepsilon_{\text{p},t}^{(i)} \\ \Delta \vect{q}_t \odot \vect{q}^{\rotwb,(i)}_t \odot \exp_\text{q} (\vepsilon_{\text{q},t}^{(i)}) \end{pmatrix}, 
                    \end{align}
                    where $\vepsilon_{\text{p},t}^{(i)}$ and $\vepsilon_{\text{q},t}^{(i)}$ are drawn from $\mathcal{N}(\vect{0},\Sigma_\text{p})$ and $\mathcal{N}(\vect{0},\Sigma_\text{q})$, respectively. 
		\STATE \textbf{Set} $t := t+1$.
  \end{algorithmic}
  \end{minipage}
\end{algorithm}

\begin{figure*}[!t]
  \centering\small
%  \pgfplotsset{
%    yticklabel style={rotate=90}, 
%    ylabel style={yshift=-15pt},
%    clip=true,
%    %colorbar,
%    colormap={parula}{
%            rgb255=(53,42,135)
%            rgb255=(15,92,221)
%            rgb255=(18,125,216)
%            rgb255=(7,156,207)
%            rgb255=(21,177,180)
%            rgb255=(89,189,140)
%            rgb255=(165,190,107)
%            rgb255=(225,185,82)
%            rgb255=(252,206,46)
%            rgb255=(249,251,14)},
%    samples=64,
%    scale only axis
%  }
  % This is the height for all of the subplots
  \setlength{\figureheight}{.28\textwidth}
  %
  % 100, 250, 480, 900, 1200, 2625
  %
  \begin{subfigure}[b]{.15\textwidth}
    \centering
   \includegraphics[]{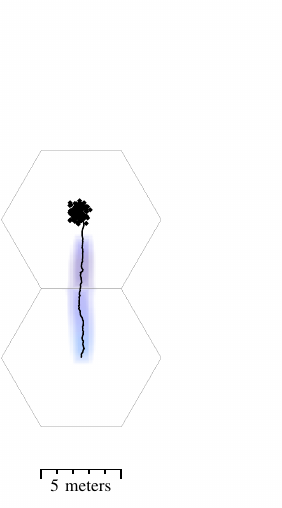}
    \caption{At $t=10$~s}
    \label{fig:dpo-a}
  \end{subfigure}  
  \hspace*{\fill}
  \begin{subfigure}[b]{.15\textwidth}
    \centering
    \includegraphics[]{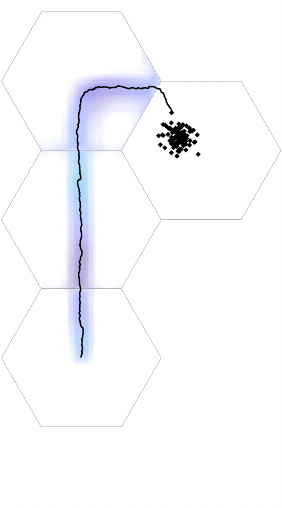}
    \caption{At $t=25$~s}
    \label{fig:dpo-b}
  \end{subfigure}  
  \hspace*{\fill}
  \begin{subfigure}[b]{.15\textwidth}
    \centering
    \includegraphics[]{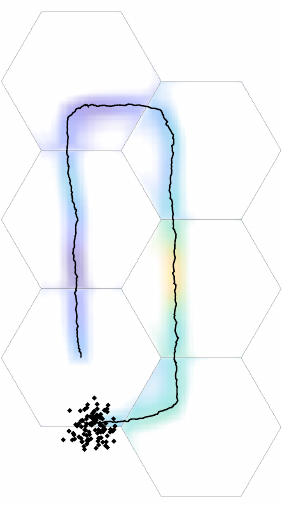}
    \caption{At $t=48$~s}
    \label{fig:dpo-c}
  \end{subfigure}  
  \hspace*{\fill}
  \begin{subfigure}[b]{.15\textwidth}
    \centering
    \includegraphics[]{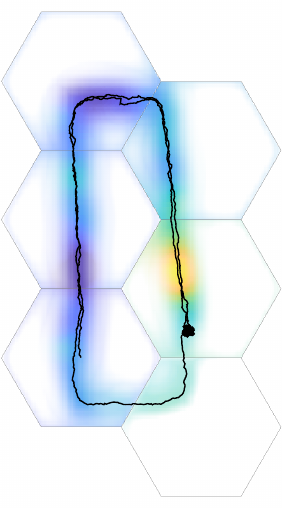}
    \caption{At $t=90$~s}
    \label{fig:dpo-d}
  \end{subfigure}  
  \hspace*{\fill}
  \begin{subfigure}[b]{.15\textwidth}
    \centering
    \includegraphics[]{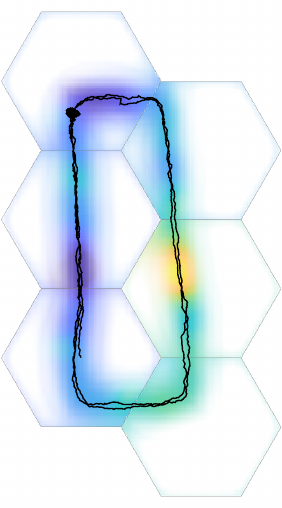}
    \caption{At $t=120$~s}
    \label{fig:dpo-e}
  \end{subfigure}
  \hspace*{\fill}
  \begin{subfigure}[b]{.15\textwidth}
    \centering
    \pgfplotsset{
      colorbar horizontal,
      colorbar sampled,
      colorbar style={font=\scriptsize, major tick length=1.5pt, 
      width=2cm, height=.33cm, at={(.5,.1)}, anchor=center},
    }
    \includegraphics[]{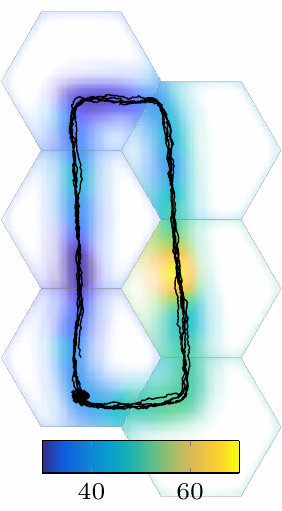}
    \caption{At $t=260$~s}
    \label{fig:dpo-f}
  \end{subfigure}
  \\
  \begin{subfigure}[b]{.17\textwidth}
    \centering
    \pgfplotsset{
      colorbar horizontal,
      colorbar sampled,
      colorbar style={font=\scriptsize, major tick length=1.5pt, 
      width=2cm, height=.33cm, at={(.5,.1)}, anchor=center},
    }
    \includegraphics[]{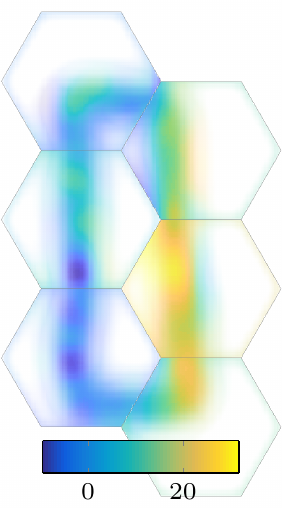}
    \caption{Magnetic $x$ ($\mu$T)}
    \label{fig:dpo-g}
  \end{subfigure}  
  \hspace*{\fill}
  \begin{subfigure}[b]{.17\textwidth}
    \centering
    \pgfplotsset{
      colorbar horizontal,
      colorbar sampled,
      colorbar style={font=\scriptsize, major tick length=1.5pt, 
      width=2cm, height=.33cm, at={(.5,.1)}, anchor=center},
    }
    \includegraphics[]{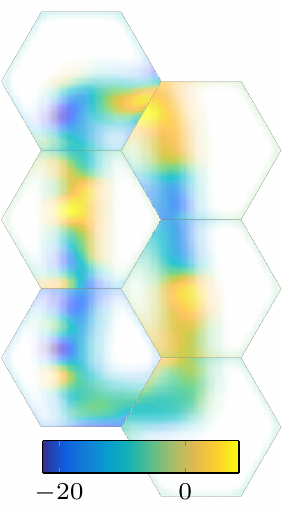}
    \caption{Magnetic $y$ ($\mu$T)}
  \end{subfigure}
  \hspace*{\fill}
  \begin{subfigure}[b]{.16\textwidth}
    \centering
    \pgfplotsset{
      colorbar horizontal,
      colorbar sampled,
      colorbar style={font=\scriptsize, major tick length=1.5pt, 
      width=2cm, height=.33cm, at={(.5,.1)}, anchor=center},
    }
    \includegraphics[]{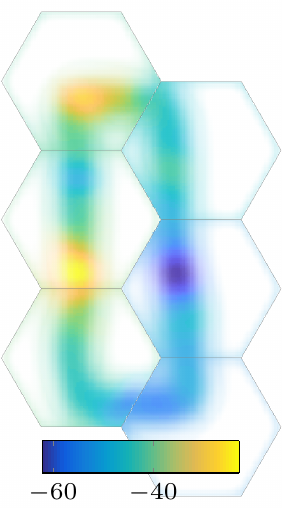}
    \caption{Magnetic $z$ ($\mu$T)}
    \label{fig:dpo-i}
  \end{subfigure}  %     
  \hspace*{\fill}
  \begin{subfigure}[b]{.14\textwidth}
    \centering
    \includegraphics[]{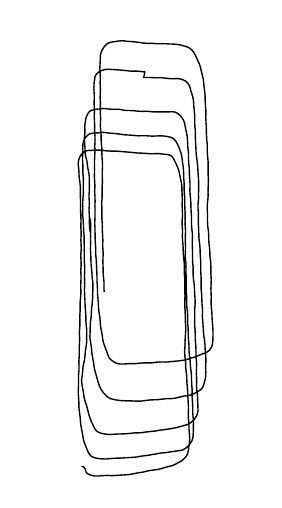}
    \caption{Odometry}
    \label{fig:dpo-j}
  \end{subfigure}  
  \hspace*{\fill}
  \begin{subfigure}[b]{.30\textwidth}
    \centering
    {\includegraphics[width=\textwidth,trim=3cm 9cm 2cm 8cm,clip]{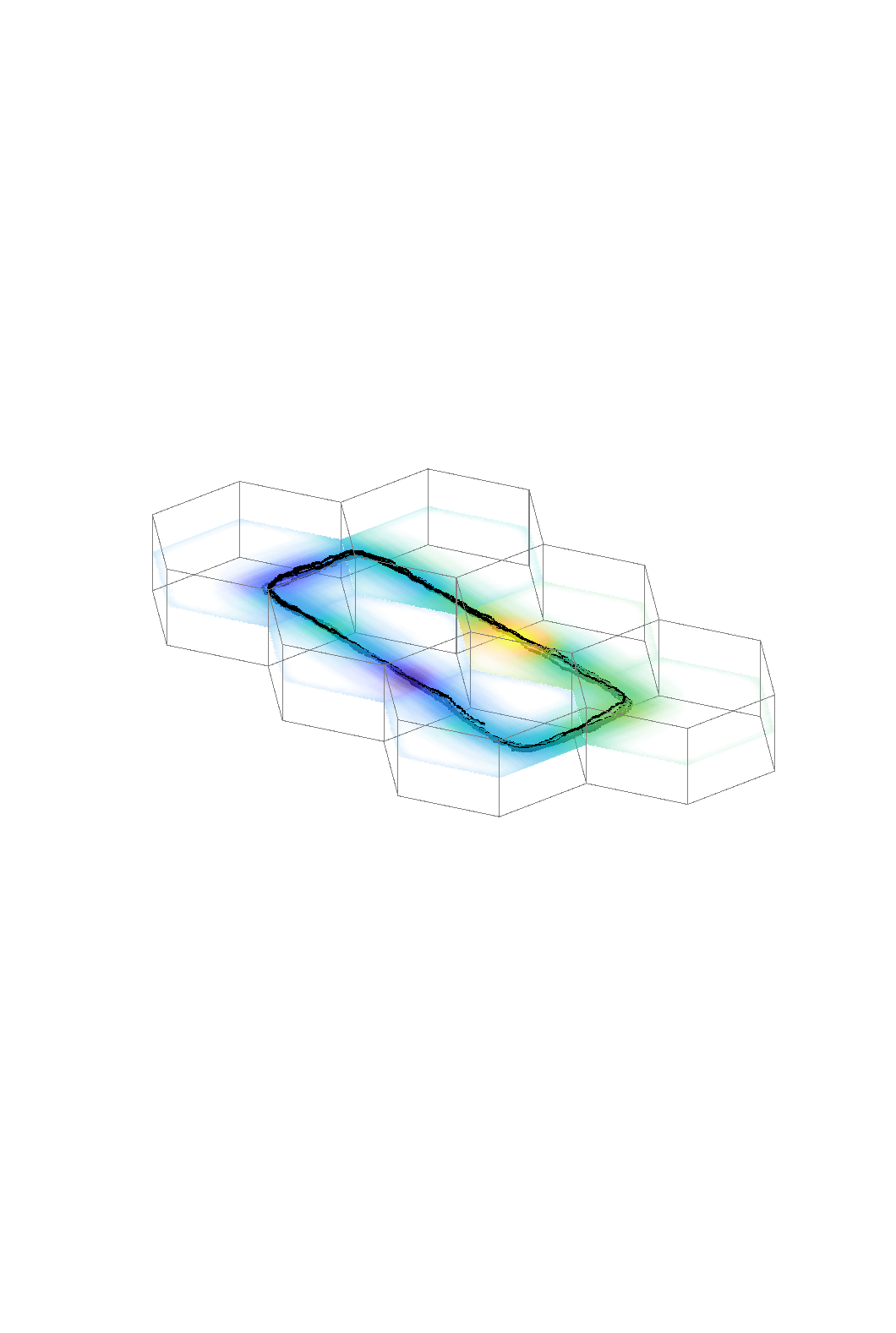}}
    \caption{Tilted 3D view}
    \label{fig:dpo-k}
  \end{subfigure}   
  \caption{The SLAM solution for walking in a square for several loops for around 250~meters. Subfigures (a)--(f) show the norm of the magnetic field map for the highest weight particle. The transparency of the maps scales with the marginal variance of the GP map. Subfigures~(g)--(i) show the corresponding vector field components for the final map in (f). The odometry trajectory is visualised in (j), and (k) shows the 3D hexagonal tiles with a cross-section of the magnetic field maps.}
  \label{fig:dpo}
\end{figure*}

\section{Results}
\label{sec:results}
\noindent
In the empirical experiments, we use the built-in magnetometer and odometry data captured by a smartphone for simultaneously building the magnetic field map and localising on it. We use an Apple iPhone~6s (published in September 2015), a previous flagship product of Apple which by current standards represents more of a standard smartphone. 

The interest in this paper is purely on building of the map and on localisation, so we choose to leverage on the recent developments in visual-inertial pedestrian dead-reckoning (PDR) methods and use the built-in ARKit PDR provided by Apple. ARKit (released September 2017, used version running in iOS~11.2.5) is using the IMU and camera of the phone to provide a 6-DoF (position and orientation) PDR movement trajectory. ARKit does visual relocalisation under the hood, which aims to correct for long-term drift, resulting in discontinuities in the PDR track. These discontinuities turned out to be more harmful than useful for our approach, and we implemented a heuristic filtering approach for removing the relocalisation jumps. The PDR---albeit being surprisingly robust---suffers from long-term drift both in position and orientation (see~Fig.~\ref{fig:dpo}).

The data was captured by a customised data capture application running on the phone. It collects the IMU (at 100~Hz) and three-axis magnetometer (100~Hz) data time-locked to the ARKit PDR estimates (60~Hz). Furthermore, the captured data was complemented with video frames of the back-facing phone camera (resolution 1280$\times$720 at 60~fps), the barometric air pressure (10~Hz), and GNSS locations captured by the phone, none of which were used in the SLAM implementation, but captured for validation and reference. All data was stored on the device, downsampled to 10~Hz before use, and the SLAM implementation was run off-line in a Mathworks \textsc{Matlab} implementation.

For simplicity and generalisability, we use the same model parameters in all our experiments. We choose the settings for the RBPF and the hexagonal blocks to be: 
\begin{itemize}
  \item We set the size of the hexagonal blocks to $r= 5~\text{m}$, $L_\mathrm{z} = 2~\text{m}$ and the number of basis functions per tile to $m = 256$. The radius and height used in solving the eigenbasis of the Laplacian extends one meter outside the hexagon.
  \item The number of particles in the RBPF is $N_P = 100$.
  \item Neighbouring tiles are updated if samples are closer than 10~cm to the hexagonal block boundary.
  \item The process noise parameters are set to $\Sigma_\text{p} = \diag(0.1^2,0.1^2,0.02^2)$ (values specified in meters drift per second) and $\Sigma_\text{q} = \diag(0.01^2, 0.01^2,0.24^2)$ (values specified in degrees drift per second).
\end{itemize}
In terms of memory usage, these choices mean that for each initialised map tile, the particle state dimension grows by 256. For fast look-up, our implementation uses a hashmap (per particle) for storing and keeping track of the map tiles. Each tile stores the GP mean and covariance matrix.

We use the same Gaussian process hyperparameter values for each of the experiments. These hyperparameters are typically learned from data. However, in sequential algorithms, hyperparameter values need to be available already before any data is processed. We therefore choose values that seem physically reasonable and that fit the data in the different scenarios that we consider reasonably well. We set the Gaussian process hyperparameters defined in Section~\ref{sec:gp-map} to $\sigma_\text{lin.}^2 = 650$, $\sigma_\text{SE}^2 = 200$, $\ell = 1.3$, and $\sigma_\text{n}^2 = 10$ (the length-scale units are meters, the magnitude parameters in $\mu$T). This assumes that the lengthscale of the magnetic field anomalies is in the order of $1.3~\text{m}$. The measurement noise standard deviation $\sigma_\text{n}$ is rather high to also account for model discrepancies. 

\subsection{Illustrating magnetic field SLAM in 2D}
\label{sec:dpo}
\noindent
In a first experiment, a data set is collected in the Design \& Project Office (DPO) at the Engineering Department of the University of Cambridge. Significant excitation in the magnetic field is present for instance due the presence of radiators and a large number of computers. The trajectory is around 250 meters long and covers an area of around $20 \times 5$~m. The same circular trajectory is traversed several times. Because of this, it is easy to visually inspect the quality of the estimates from our SLAM algorithm which are shown in Fig.~\ref{fig:dpo}. The different subfigures illustrate the workings of our algorithm. As a supplementary file to this paper, there is also a video\footnote{The supplementary video is available on YouTube: \mbox{\url{https://youtu.be/pbwWLoh6mvI}}} demonstrating the workings of the SLAM algorithm. 

Figs.~\ref{fig:dpo}(a)~--~\ref{fig:dpo}(f) show how the estimates progress over time. For visualisation purposes, the norm of the magnetic field is predicted by the map of the highest weight particle at a large number of locations in each hexagon. The transparency in the map visualises the uncertainty of the map. It can be seen that the uncertainty is large in unexplored regions (no map is actually visible) and that the uncertainty of the map decreases over time in the re-visited regions. 

The black line displays the trajectory of the highest weight particle. In Figs.~\ref{fig:dpo}(a) and (b), the deliberate slight delay in updating the map can clearly be seen. As discussed in Section~\ref{sec:rbpf}, we wait with resampling until at least 90\% of the particles is in a tile that they have previously updated and left. Because of this, no resampling is done in the first loop, until around 50 seconds and the spread in the particles in Fig.~\ref{fig:dpo}(c) is therefore fairly large. This allows the algorithm to properly close the loop as visualised in Fig.~\ref{fig:dpo}(d). Another aspect to note about our algorithm is that in Fig.~\ref{fig:dpo}(c), most particles are on the border between the tile in which the trajectory was started and a new tile below. To allow for smooth transitions between these two tiles, many of the particles will in this case  update the local magnetic field maps in both tiles. 

The map that is visualised in Figs. \ref{fig:dpo}(a)~--~(f), shows the norm of the predicted magnetic field. Our model~\eqref{eq:scalar-potential-gp}, however, explicitly models the magnetic field as a three-dimensional vector. Instead of visualising the norm of the predicted magnetic field, it is therefore also possible to visualise the predicted magnetic field in three directions. This is shown in Figs.~\ref{fig:dpo}(g)~--~(i). Because the dip angle of the Earth's magnetic field is fairly large in Cambridge, UK, the $z$-component of the magnetic field is of significantly larger magnitude than the $x$- and $y$-components. 

Based on Fig.~\ref{fig:dpo}(f) it can be concluded that our SLAM algorithm estimates the trajectory with high accuracy since the same trajectory can be seen to be traversed multiple times. For comparison, we also show the odometry in Fig.~\ref{fig:dpo}(j) which clearly shows a drift in the position. Our magnetic SLAM approach is able to correct for this drift. To illustrate the fact that even though the movement for this data set is in a 2D plane the estimation is still done in 3D, the tilted 3D view is shown in Fig.~\ref{fig:dpo}(k). 

\begin{figure}[!t]
  \centering\small
  {\includegraphics[width=0.45\columnwidth,trim=3cm 8.5cm 2cm 5cm,clip]{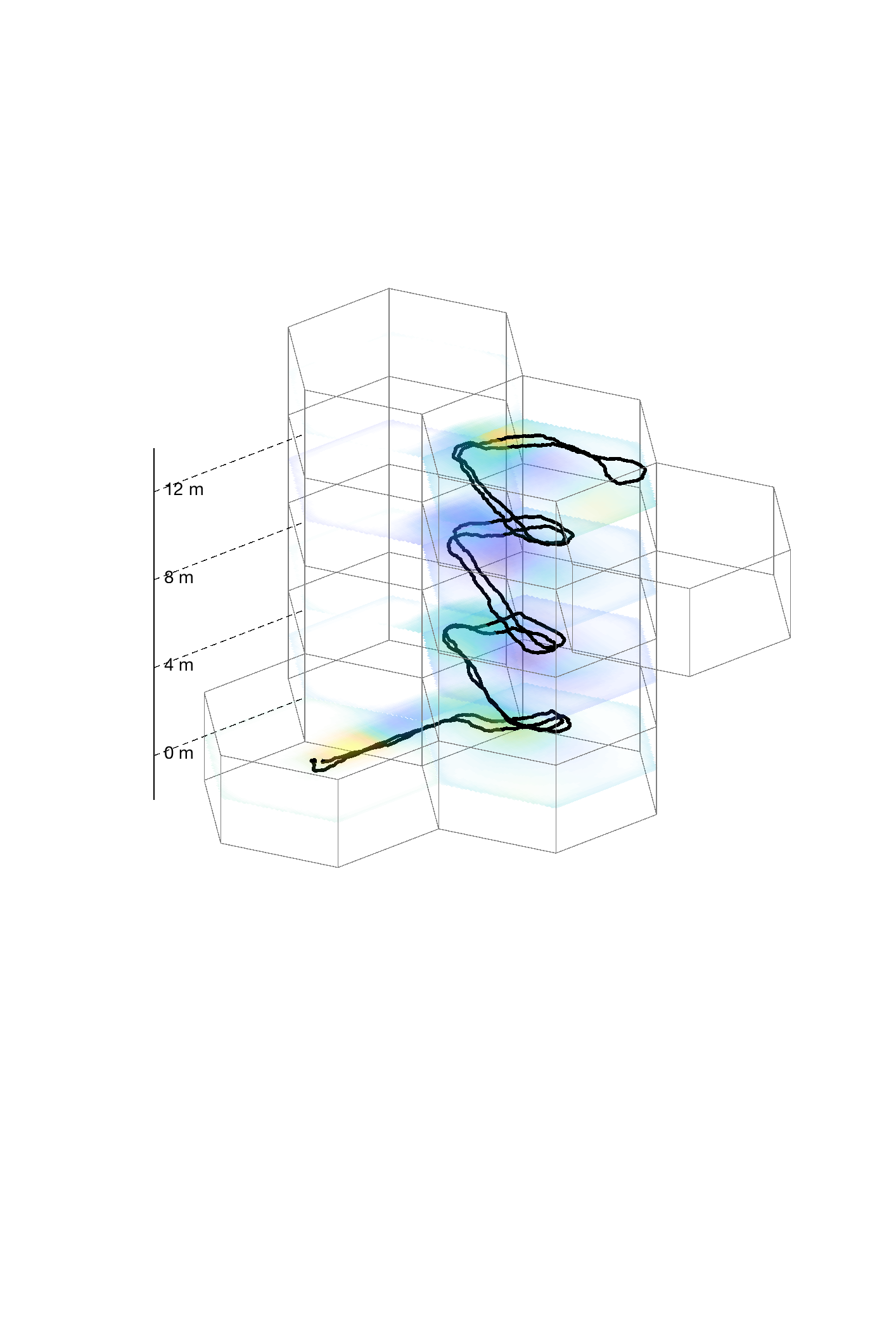}}
  \caption{Example showing the three-dimensional nature of our SLAM algorithm. The path covers a flight of stairs in the University of Cambridge Engineering Department. The complete path length is 125~meters. The cross-sections of the tiles show the magnetic map norm (the color scale is the same as in Fig.~\ref{fig:dpo}(f)).}
  \label{fig:staircase}
\end{figure}

\subsection{Fully 3D magnetic field SLAM}
\noindent
In Section~\ref{sec:dpo}, the movement was close to two-dimensional. However, exactly the same procedure also applies to movement and block tile structures in three dimensions. In Fig.~\ref{fig:staircase} we show the estimated trajectory and the magnetic field map for an empirical experiment in which we first walked up and subsequently down a set of stairs at the Engineering Department of the University of Cambridge. 

Again, each hexagonal block has its own local magnetic field map. To visualise these, the norm of the predicted magnetic field for a hexagonal slice in the middle of each block is shown. The transparency visualises the uncertainty of the map. Just as in the results presented in Section~\ref{sec:dpo}, our algorithm only starts to resample particles on the way down the stairs. This results in an estimated path that is fairly similar in both directions. 

\subsection{Large-scale magnetic field SLAM}
\begin{figure}[!t]
  \centering\small
  \pgfplotsset{
    yticklabel style={rotate=90}, 
    ylabel style={yshift=-15pt},
    clip=true,
    %colorbar,
    colormap={parula}{
            rgb255=(53,42,135)
            rgb255=(15,92,221)
            rgb255=(18,125,216)
            rgb255=(7,156,207)
            rgb255=(21,177,180)
            rgb255=(89,189,140)
            rgb255=(165,190,107)
            rgb255=(225,185,82)
            rgb255=(252,206,46)
            rgb255=(249,251,14)},
    samples=64,
    scale only axis
  }
  % This is the height for all of the subplots
  \setlength{\figureheight}{.5\textwidth}
    \pgfplotsset{
      colorbar horizontal,
      colorbar sampled,
      colorbar style={font=\scriptsize, major tick length=1.5pt, 
      width=3cm, height=.33cm, at={(0.75,.1)}, anchor=center},
    }
    \includegraphics[height=0.5\textwidth]{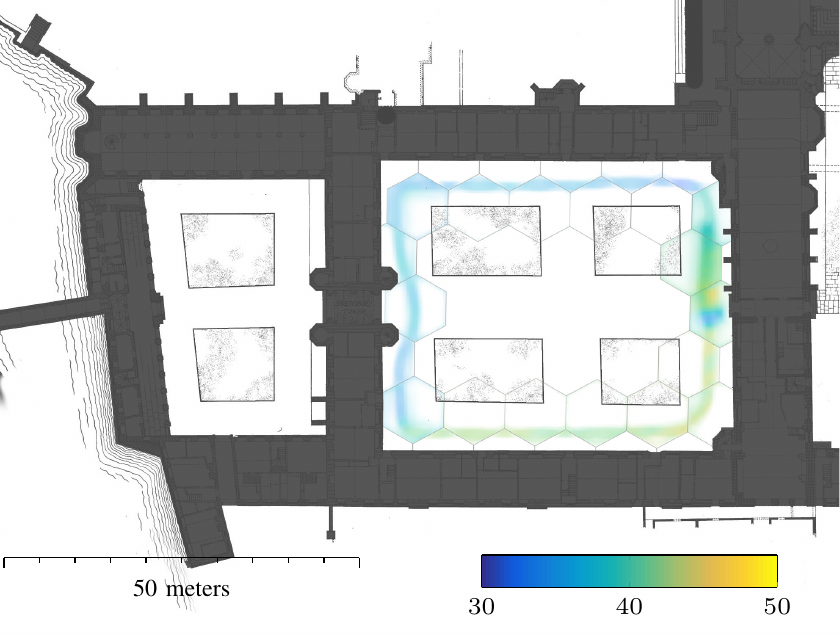}
  \caption{The SLAM map solution for a large-scale outdoor--indoor example. The visualised color field is the norm of the magnetic field, and the opacity follows the uncertainty. The layout corresponds to that of Fig.~\ref{fig:intro}.}
  \label{fig:st-johns}
\end{figure}

\noindent
In a third empirical data set we walk around St.\ John's college in Cambridge. The college consists of several courts and even though the data set is captured outdoors, the close proximity of the buildings results in small magnetic field anomalies that can be used for localisation. The SLAM results were previously presented in Fig.~\ref{fig:intro} and are presented in more detail in Fig.~\ref{fig:st-johns}. The map extends over 27 tiles covering the courtyard. For many algorithms the extent of the data set would therefore be prohibitively large. However, due to our computationally efficient model in terms of hexagonal tiling, our approximate Gaussian process model and the RBPF algorithm, the computations remain feasible. It can therefore be concluded from Fig.~\ref{fig:st-johns} that our approach scales to large-scale magnetic field SLAM. 

\section{Conclusions and future work}
\noindent
We have presented a scalable algorithm for magnetic field SLAM in 3D. It builds local Gaussian process maps where each map is an approximation to the local full GP solution in terms of a basis function expansion. This approximation allows us to map a magnetic map volume of $\sim 260~\text{m}^3$ in a state dimension of only 256. Each local map is modelled in terms of a hexagonal block tile, which minimises the number of tiles that are needed to cover the total mapping volume. The formulation fits perfectly in an RBPF resulting in a computationally tractable algorithm. The computational complexity scales linearly both with the number of particles in the particle filter and with the number of hexagonal tiles of each particle. The algorithm is shown to perform well on three challenging empirical data sets.

One of the challenges in the data we used for testing, is that the odometry we obtain from the ARKit does not always obey our motion model. More specifically, for straight paths the odometry shows almost no drift, while fast turns can introduce sudden errors. This does not follow our motion model in which we model a constant process noise. In future work we would like to focus on using inertial PDR as odometry instead. Apart from avoiding these issues, this would also result in a positioning algorithm only based on widely available inertial and magnetometer sensors. Other directions of future work are in the direction of using a Rao--Blackwellised Particle Smoother and further reducing the computational complexity of the algorithm. 

\section*{Acknowledgments}
\noindent
This research was financially supported by the EPSRC grant \emph{Autonomous behaviour and learning in an uncertain world} (Grant number: EP/J012300/1) and the Academy of Finland grant \emph{Sequential inference for real-time probabilistic modelling} (Grant number: 308640).

%

%\IEEEtriggeratref{17}

% References section
%\bibliographystyle{IEEEtran}
%\bibliography{IEEEabrv,bibliography}

\bibliographystyle{unsrtnat}
\bibliography{bibliography}

\end{document}